\documentclass[twoside]{article}

\usepackage[accepted]{aistats2019}
\usepackage{graphicx}

\begin{document}

%

%

\twocolumn[

\aistatstitle{Cooperative Multi-Agent Reinforcement Learning Framework for Scalping Trading}

\aistatsauthor{Uk Jo \And Taehyun Jo \And Wanjun Kim}
\aistatsaddress{ModuLabs, Seoul, KR}
\aistatsauthor{Iljoo Yoon\And Dongseok Lee \And Seungho Lee}
\aistatsaddress{ModuLabs, Seoul, KR}]

\begin{abstract}
We explore deep Reinforcement Learning (RL) algorithms for scalping trading and knew that there is no appropriate trading gym and agent examples. Thus we propose gym and agent like Open AI gym in finance. Not only that, we introduce new RL framework based on our hybrid algorithm which leverages between supervised learning and RL algorithm and uses meaningful observations such order book and settlement data  from experience watching scalper’s trading. That is very crucial information for trader’s behavior to be decided. To feed these data into our model, we  use spatio-temporal convolution layer, called Conv3D  for order book data and temporal CNN, called Conv1D for settlement data.  Those  are preprocessed by episode filter we developed.  Agent consists of four sub agents divided to clarify their own goal to make best decision.  Also, we adopted value and policy based algorithm to our framework. With these features, we could make agent mimic scalpers as much as possible. In many fields, RL algorithm has already begun to transcend human capabilities in many domains. This approach could be a starting point to beat human in the financial stock market, too and be a good reference for anyone who wants to design RL algorithm in real world domain. Finally, we experiment our framework and gave you experiment progress.
\end{abstract}

\section{Introduction}

Nowadays, RL has recently been introduced and applied to solve challenging not only Game~\cite{Mniha}, but also real world problems~\cite{Li2015}. But in finance RL is so challenging to adapt. First, it is very hard to proceed under the assumption that the environment itself follows Markov Decision Process and the roles of the agents are mixed up and not clear to achieve a goal. 
Second, buy and sell agent needs to have different observation and reward. But many studies are done by a single agent. Therefore, rather than being given a clear goal, the RL agent itself has a disadvantage in that it performs multiple missions like a multi-agent in a single agent.
Third, research in most financial markets has often led to stock price forecasting through simple learning and showing associativity is not helpful in practice. The absence of trading algorithms that have gained the performance of RL, which is rapidly evolving, has also become the background to this study
Lastly, the success or failure of investment decisions does not end with simply predicting stock prices, but rather depends on the degree of confidence in the stock price prediction

To overcome the shortcomings of this study, we focused on three things. Data, episode design, framework, and algorithms.

\subsection{Data}
It do not use the price information provided by the simple candle charts for the second trade, but rather used the original data of the transaction, the ask and the closing data. The price, price, price, and closing price information provided by the candle chart are data that are direct evidence of price formation. It is assumed that the data is the lowest level in which transactions are actually concluded and prices are formed, and that the utilization of such data is suitable for learning of short-term trade algorithms.

\subsection{Episode design}
It was started watching scalper's trading habit and environment. These traders are traded on the order book and transaction information within a short period of time. This is a very good environment for RL. RL could catch traders' desire and habit to maximize their return. As a result, we decided to process the tick data, which is a much smaller range, so as to learn the transactions per second. In most cases, the short-term transaction is more likely to satisfy the Markov assumption, which is the background of RL, because it is unlikely that external variables such as changes in company value or adverse or negative events will be affected. It is assumed that the endless desire of many people participating in the stock market will generate revenue generating patterns repeatedly, and that RL algorithms will outperform any other algorithms in this pattern.

\subsection{Framework}
There are Episode Filter, Gym, and Agent. We developed an agent to mimic the behaviors and selection of short - term investors in four agent - design cooperating environment and reward - definite environment similar to episode definition and trading system.

First, Episode filter, which is a learning environment, is a real-time API that facilitates easier experimentation by easily filtering and scaling ticks data collected during April for stocks that are suitable for short-term trading. This means that short-term traders choose the stocks that match their conditions or criteria that they can trade well and concentrate on trading the stocks, so we can use the same type of stocks (for example, And K-times increase in trading volume), we developed the system to extract only the stock prices of specific conditions in order to learn the RL agent specialized in the stock price pattern.

Second, Trading Gym was developed in accordance with OpenAI's Gym interface, which provides a game environment for learning reinforcement learning algorithms. This was developed to provide state and observation to the reinforcement learning agent and to interact with the action of the reinforcement learning agent. This trading gym makes it easier to calculate rewards and returns based on actual transaction taxes and transaction fees.

\subsection{Algorithms}
First, unlike other trading RL algorithms, Obtain information of observation through Spatio-temporal CNN. Real traders do not know and memorize all the explicit numbers when they are short-term traded, but execute the trading by grasping the overall flow and characteristics rather than executing the trading. Likewise, I thought that applying CNN to reinforcement learning agents would mimic them.

Second, in order to solve a problem that is difficult to solve, it is possible to effectively solve a difficult problem when one problem is divided into several different problems. For example, if a non-walking robot first learns to "move in a certain direction" and then learns to navigate the maze in order to solve the problem of finding a space within a short distance, Correctly correct the problem~\cite{Frans}. Likewise, the study of stock trading also takes a long learning time and it would be difficult to learn properly if the learning is provided by simply providing a reward of one agent as a return rate. Therefore, in this study, the agent has to divide into four major tasks to do the actual stock trading, so that it can concentrate on each role, and it is configured to achieve excellent overall performance.

Third, the success or failure of the actual investment decision does not end with simply predicting the stock price, but depends greatly on the degree of confidence in the stock price prediction. Therefore, we adopted a learning method that imitates traders that maximize returns based on the accuracy of the predictions, and used RL to maximize returns and outcomes of guidance learning to determine the magnitude of the predictions.

\section{Related Works}
In the past, research on patterns of financial markets was more macroscopic. There was a lot of effort to grasp the aspect of the market from at least one day data. Microscopic behavior of the market was understood as noise. In recent years, however, research has been carried out that reveals that the behavior of micro-markets is quite meaningful~\cite{Madhavan2000}. The quotation and closing data we used are microstructure data in the market. These data show how the price of financial assets is formed and in what pattern. Market microstructure data show price formation and price discovery at the most fundamental level~\cite{norges_report}.
We believe that by utilizing these market microstructure data, we can grasp the repetitive behaviors of the market participants and get the opportunity of excess return or absolute profit in the market. Also, RL needs to understand the domain to adapt and agent is designed by experience of pro game player or well-trained operator, so called expert. ~\cite{VinyalsTimoEwaldsSergeyBartunovPetkoGeorgievAlexanderSashaVezhnevetsMichelleYeoAlirezaMakhzaniHeinrich}  Thus we watched that traders trade in stocks with fluctuating price fluctuations with price information and concurrence amount of one to two minutes. As far as we know, there are not many cases of studying with such data and exchange. It was possible to predict the actual price with the window information. ~\cite{Han2015}, which means that there is a significant correlation between the order book and transaction information and the future price. Also, to extract state of agent, we use Conv3D  for order book and Conv1D for transaction observation~\cite{Diba,UlVarol}. We assumed that in endless episode CNN variants are better than RNN~\cite{Hausknecht}. To make agent mimic traders, we also prepared data of order book and transaction data.~\cite{Jiang2017}.
Clarify we adopt Four independent agents were used to increase performance~\cite{Lee2007}
In the case of AlphaGo, we used the notes of experts in the game to lead learning first and second learning through self-play. This is because, if there is label data to learn, it is possible to make a better agent faster by learning RL after learning Q network in RL~\cite{Silvera,Jin}. Through this, we trained the Q network by using the collected data through supervised learning and then we could complement this value with Q value with  another Q value from RL. It is inspired by the multi-agent learning using the global q value by summing the Q values obtained from each agent~\cite{Rashid2018}. We adapt this mixed Q network into Asynchronous Actor Critic(A3C)~\cite{Mnih}, DQN~\cite{Mniha}.  Like ~\cite{Silver}, we decided to implement our framework and algorithms as a whole system and has plan to publish for researchers.

 As we know, most of the stock market papers have seen that the training data and test data are separated from each other and only the test data is used to show the results by staying in the training phase rather than verifying the performance.

\section{Proposed Frameworks}

\subsection{Overall Concepts}

The data required for learning and verification are the stock prices and contract data of the KOSPI and KOSDAQ stocks from April to July 2018 in the Korean stock market. The data were filtered using the Episode Filter, which showed a 15\% increase over the previous day. Of these data, 70\% of the data were randomly selected as Train Data and 30\% as Test Data. These data were used for learning by preprocessing.

The network of each agent learns through two steps of guidance learning and RL~\cite{Silver,Silvera}.

\subsubsection{Supervised Learning}
RL is generally slower than instructional learning because it learns with a reward experienced through random behavior for many states. In particular, unlike other data, stock data is not determined by the state alone, but is slower because of the strong random-walk property. In addition, since the four agents share mutual learning and share the reward, there is a problem that the variance becomes too large because severe noise is generated in the reward due to the behavior of other agents that are not learning at the beginning of learning.
To solve this problem, pre-training was performed through each map learning before four agents learned mutually. This allows agents to learn quickly in advance, greatly reducing learning time. In addition, it greatly reduces the noise of input and reward of other agents caused by untrained agents at the beginning of mutual learning, so that the policies of other agents converge well.

\subsubsection{Reinforcement Learning}
When the policies of the agents are converged to some degree through map learning, the agents are connected to construct the whole network and learn by using DQN. Agent learned the results of the map learning by linking it with the learning network which has stopped the learning so that it can learn sufficiently through RL, and the RL network which has such structure but can learn.
After the agent completes each process, it will then receive an additional reward by multiplying the agent's reward by a certain percentage. This sharing of rewards enables not only the role of each but also the whole agent to learn in the direction of maximizing profit.

\subsection{Architecture}
The RL agent that learns stock trading is composed of four agents having different roles needed for stock trading, not a single agent. Each agent has a reward appropriate to its role, and the secondary reward linked to the performance (return) of the whole agent is added to the reward. 

\begin{figure}[h]
\vspace{.3in}
\centerline{\fbox{\includegraphics[width=1\linewidth]{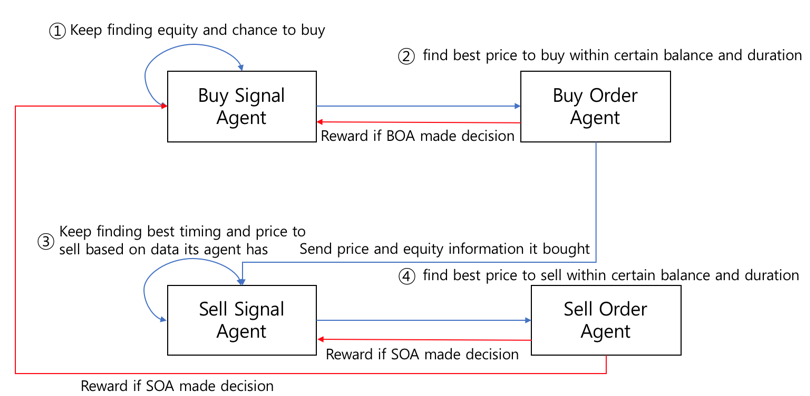}}}
\vspace{.3in}
\caption{Overall Agents Architecture}
\end{figure}

\paragraph{Buy Signal Agent(BSA)} Based on the observation, BSA predicts the time ($t_1$) at which the share price is expected to rise steadily for two minutes from the present point in time.

\paragraph{Buy Order Agent(BOA)} After the time ($t_1$) predicted by the BSA, the BOA proceeds with the stock purchase based on the Observation information. At this time, the BOA will buy the shares at the lowest possible price (buy time $t_2$).

\paragraph{Sell Signal Agent(SSA)} After the BOA completes the buying ($t_2$), the SSA predicts the time ($t_3$) at which the stock price is expected to fall for two minutes from the present time, based on the Observation information.

\paragraph{Sell Order Agent(SOA)} After the SSA predicted time ($t_3$), the SOA proceeds with the sale of shares based on the Observation information. At this time, the BOA sells the stocks at the most cost (buy time $t_4$).

\paragraph{Agent Common constraints} All agents have a first order reward estimation method for each action so that they can learn their roles. In addition to this, you will receive a second reward by multiplying the reward of the remaining agent except the agent by 0.5 times. Through this, they learn the role of each agent (first reward) while learning the role of all agents (second reward).
All agents have time limits from the time ($t_1$) predicted by the BSA to the time $t_1 + 120$. If all agents' actions are not completed until $t_1 + 120$, all agent actions are enforced at $t_1 + 120$.

\subsection{Data Preprocessing}

We collected all the tick data of KOSPI and KOSDAQ (Korea Stock Exchange Market) for about four months for agent learning. This data includes the order book data and price and trade volume for all time points.
Depending on the stock, the price and the trade volume of each stock are very different. Also, even if they have the same numerical value, the numerical value has different meaning. For example, one share of Samsung Electronics(005930) and KCS(115500) would be very different in regard of meaning of the degree of influence of price. We have to normalize the prices and shares of each stock to distinguish these differences for RL agents. It is necessary to scale data in order for the RL agents to learn from different numerical values on the same viewpoint.

\paragraph{Price} can be normalized based on the price at a specific point in time. Since our study considers short-term trading, it would be advantageous to learn that the nearest price was the standard. So, we normalized based on the previous day's closing price.

$S_{p}$ \hspace{1cm} scaling function

$P_t$  \hspace{1cm} price at time $t$

$P_y$  \hspace{1cm} close price at yesterday

\begin{equation}
S_{p}(P_t)=\frac{P_t - P_y}{P_y} \times 100
\end{equation}

\paragraph{Shares}
Outstanding shares among each stock are very different. Trading volume would be normalized based on the number of shares outstanding. More precisely, We use the stake of the majority shareholder on each stock to consider how many shares are in the market. These values are converted into a logarithmic scale so that the overall values are within a certain range.

$S_{\textrm{shares}}$ \hspace{1cm} function share for scaling

$V_t$ volume at time $t$

$Sh_{\textrm{outstanding}}$  number of shares outstanding

$Sh_{\textrm{majority}}$  number of shares owned by major

\begin{equation}
S_{\textrm{shares}}(V_t) = \ln \left(\frac{V_t}{Sh_{\textrm{outstanding}} - Sh_{\textrm{majority}}} \right)
\end{equation}

\subsection{Observation}
Agents will receive 120 pieces of information (120 seconds) in 51 seconds, including current second.
\begin{table}[h]
\caption{Observations} \label{tab:oversavations}
\begin{tabular}{ll}
\hline
                    & Observations                                                                                                                                                                                                                                                                                                                                               \\ \hline
\textbf{Ask/Bid}    & \begin{tabular}[c]{@{}l@{}}Bid Price  1 $\sim$10\\ Ask Price 1 $\sim$10\\ Bid Amount 1 $\sim$10\\ Ask Amount 1 $\sim$10\end{tabular}                                                                                                                                                                                                                       \\ \hline
\textbf{Trading Volume} & \begin{tabular}[c]{@{}l@{}}Last Price,\\ Trading Volume,\\ Sell direction Volume,\\ Weighted avg sell price,\\ Buy direction volume \\ Weighted avg buy price,\\ Total directional volume,\\ Total direction weighted \\ avg price, \\ Open price on the day,\\ High price on the day,\\ Low price on the day\end{tabular}
\end{tabular}
\end{table}

Each of these basic Observations is additionally provided with information necessary for trading.

$O_t$  \hspace{1.3cm} observation at time $t$

$LT_t$  \hspace{1.1cm} remaining time at time $t$

\begin{table}[h]
\caption{Agents' Observation} \label{tab:agents_obs}
\centering
\begin{tabular}{cc}
\hline
\textbf{Agent} & \textbf{Observation} \\ \hline
BSA            & $O_t$                    \\
BOA            & $O_t$, $LT_t$                \\
SSA            & $O_t$, $LT_t$               \\
SOA            & $O_t$, $LT_t$              
\end{tabular}
\end{table}

\subsection{Rewards}

Each agent receives a primary reward depending on how well he or she has performed his role. After that, you will receive additional secondary rewards depending on how well the remaining agents have performed the role since each agent.
Secondary rewards will receive an additional 0.5 times the rewards the remaining agents will receive after this process for each agent.
Each agent learns to perform his role as best as possible through the primary reward. However, in order to achieve a good performance in actual investment decisions, it is necessary to make judgment on buying, real buying, selling judgment, and actual selling. Secondary rewards are used to learn not only the role of each agent.
Basically, each agent counts reward as 0 if it did not take an action at each step, and if it took an action, it gets reward according to action for each agent. The reward according to agent action is as follows.

\paragraph{BSA} 
If the BSA generates a signal, the average rate of increase for the next two minutes is rewarded based on the price ($P_{t_{1}}$) at which the BSA generated the signal ($t_1$). That is, the more stable the stock price rises for two minutes from the starting point, the better the reward will be.
\begin{equation}
    R(t=t_{1} | a=1) = \frac{1}{120} \sum_{t=t_{1} + 1}^{t_{1} + 120} \frac{P_{t} - P_{t_{1}}}{P_{t_{1}}} \times 100.
\end{equation}

\paragraph{BOA} 
After the BSA has generated the signal ($t_{1}$), you will be rewarded how cheaply the BOA has made the purchase, compared to the lowest price before the buyout. At this time, the point at which the BOA completes the purchase is called.
\begin{equation}
    R(t=t_{2} | a=1) = \frac{P_{t_{2}} - \min (P_{t_{1}}, P_{t_{1}+1}, \cdots, P_{t_{2}})}{\min (P_{t_{1}}, P_{t_{1}+1}, \cdots, P_{t_{2}})} \times 100. 
\end{equation}

\paragraph{SSA}
Based on the time ($t_{3}$) at which the SSA generates the signal, the mean value of the absolute value of the decline rate for the remaining time is received as a reward. As the SSA signals, the stock price falls a lot and you get a good reward. That is, if it is judged that a lot of stocks will be dropped in the future, the learning proceeds toward the signal generating side.

\begin{equation}
    R(t=t_{3} | a=1) = \frac{1}{LT_{t_{3}}} \sum_{t=t_{3}+1}^{t_{3}+LT_{t_{3}}} \frac{-(p_{t} - P_{t_{3}})}{P_{t_{3}}} \times 100,
\end{equation}
where $LT_{t}$ is remaining time at time $t$.

\paragraph{SOA}
After the four agents have completed the transaction, they will receive the actual return rate as a reward. That is, the price ($P_{t_{4}}$) sold by the SOA to the price ($P_{t_{2}}$) bought by the BOA is calculated as the return rate and rewarded.

\begin{equation}
    R(t=t_{4} | a=1) = \frac{P_{t_{4}} - P_{t_{2}}}{P_{t_{2}}} \times 100.
\end{equation}

\subsection{Actions}
After the four agents have completed the transaction, they will receive the actual return rate as a reward. That is, the price () sold by the SOA to the price () bought by the BOA is calculated as the return rate and rewarded.

\begin{table}[h]
\caption{Agents' Actions} \label{tab:agents_actions}
\centering
\begin{tabular}{@{}ll@{}}
\textbf{Agents} & \textbf{Actions}             \\
\hline
BSA             & signal(a=1), not signal(a=0) \\
BOA             & buy(a=1), not buy(a=0)       \\
SSA             & signal(a=1), not signal(a=0) \\
SOA             & sell(a=1), not sell(a=0)    
\end{tabular}
\end{table}

\subsection{Network}

The flow of the whole transaction is as follows
\begin{figure}[h]
\vspace{.3in}
\centerline{\fbox{\includegraphics[width=1\linewidth]{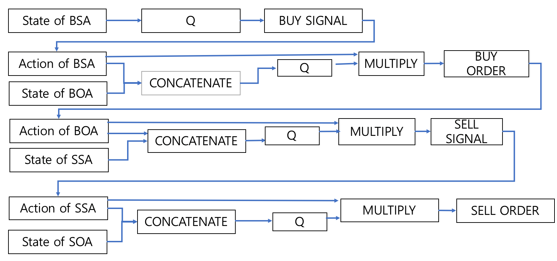}}}
\vspace{.3in}
\caption{Reinforcement Agents Structure}
\end{figure}

Rather than simply accepting observation information as explicit information, I thought it was easy for learning to accept information that was extracted through CNN Network. CNN thought it was the same way that real-world traders focused on the overall flow of data rather than focusing on each number of quotations and closing data when making investment decisions.
Each agent is a CNN network that can process SpatioTemporal information composed of three-dimensional data as follows ~\cite{Diba,UlVarol}.

\begin{figure}[h]
\vspace{.3in}
\centerline{\fbox{\includegraphics[width=1\linewidth]{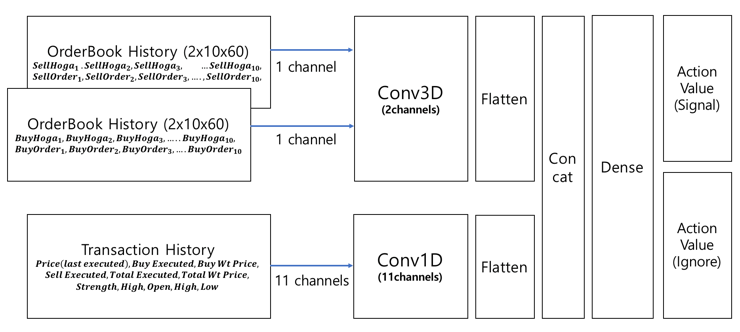}}}
\vspace{.3in}
\caption{CNN Structure}
\end{figure}

We used two convolutional layers to collect the time series data and to extract the features.
Order-Book History used the Conv3D Layer to extract the temporal relationships between features in a single call window and multiple call windows. In this case, the buy and sell prices are data having different characteristics, so they are processed using two Conv3D layers.
Transaction History has time-series characteristics of each data, but since there is no correlation between each data, only the time-series characteristic is extracted using Conv1D layer with 11 channels.
Data passing through the convolutional layer passes through the flattened layer, passes through the dense layer, and finally judges the Order / Not Order.

\subsection{Train Method}
\label{sec:train_method}

Each agent's network learns through two levels of guidance learning and RL [8, 9].

Since the stock data itself assumes that the action of the agent did not affect the agent, the agent responsible for the first part of the transaction sequence, such as the Buy Signal Agent, can quickly learn by learning the map before linking the four agents.

The Signal Agent calculates the area of the threshold 2\% for the future price for each Observation, and learns the Observation with the input and the area with the output. Then, the value of Q value is converged through RL in which the area value is rewarded.

The Order Agent uses a random signal along with each Observation, and conducts a map learning with the signal price and transaction price as output values. Then, the value of Q value is converged through RL in the same way.

When the policies of the agents are converged to some extent, the agents are connected as shown in the above figure to configure the whole network and learn using DDQN. For each transaction sequence, the agent learns its own state by linking the agent action of the previous stage.

\section{Experiments}

Section~\ref{sec:train_method} As mentioned in the Train Method, experiments are largely divided into instructional learning and RL. After learning the map, the agent completes some learning quickly for each role, and all four agents are connected to complete mutual RL.

\paragraph{ Map learning}
Map learning was conducted using CNN network for each agent. We experimented with changing parameter sets to find optimal network parameters for each agent.
The number of Epoch, Batch size, and Neuron was controlled by the change of the parameter set. Each figure attached to the agent at the bottom shows the result of plotting the MAE, MAPE, Theil's U, and Correlation values ​​of the best-learned network based on the loss in the agent's map learning according to the epoch.
The first table shows the best combination of five network parameters (Epochs, Batch size, Neurons) when learning by changing parameter sets. The second table is MAE, MAPE, Theil's U, and Correlation values ​​after the learning of the five networks with the best learning.
After the learning of the map, the learned CNN network was transplanted into the RL network and set as the weight value of the initial network of RL. Through this process, each agent started roughly with a network weight for his role, which saved a lot of learning time.

\paragraph{ Reinforcement Learning}

Through the learning of maps, each agent learned a certain amount of network weight for each role. However, it is important that the four agents perform their respective roles well, but ultimately, the final goal is to learn transactions that maximize returns in a harmonious manner.
In the RL stage, as described in Section 4.7, learning about the role of each agent is similar to that in the map learning through the first reward. However, by receiving the second reward according to the result of the final sale beyond each role, We also learn about the harmonious role of agents.
The experimental results of reinforced learning are as follows.

\begin{figure}[h]
\vspace{.3in}
\includegraphics[width=1\linewidth]{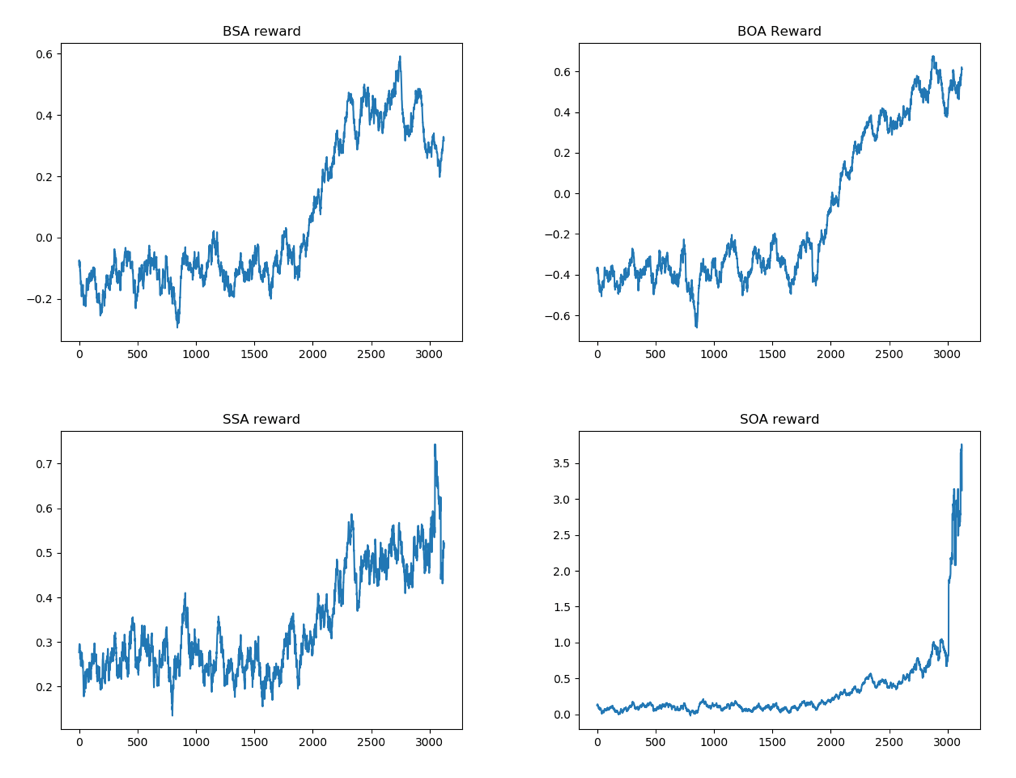}
\vspace{.3in}
\caption{Agents reward}
\end{figure}

\paragraph{Agents Reward}

Reward showed a rise over 2000 episodes as the episode progressed. The BSA, which existed at the beginning of the transaction and received the least influence from other agents, converged first, and the SOA that received the greatest influence from other agents was the last to converge.

Unlike games like Atari, stock data is very sensitive to overfitting. Therefore, we decided whether to start trading and decide the timing of ending the learning based on the BSA which has the greatest effect on the profit rate.

\paragraph{Episode average Profit}

\begin{figure}[h]
\vspace{.3in}
\includegraphics[width=1\linewidth]{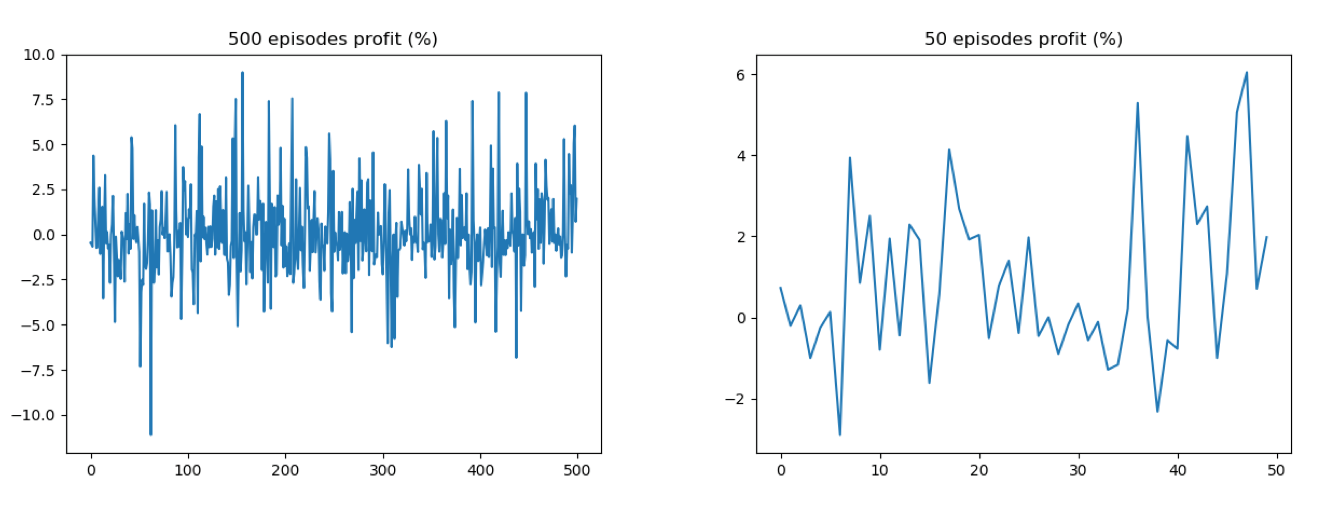}
\vspace{.3in}
\caption{Episode average profit}
\end{figure}

The left side shows the average of the sum of the last 500 episodes, and the right side shows the average of the last 50 episodes.

The average of the 500 episode returns is 0.076\%, but the average of 50 episode returns which all agents have converged enough is 0.807\%.

\begin{table}
\caption{BSA Training Top 5 Parameters Table} \label{bsatable}
\centering
\begin{tabular}{lllll}
\textbf{No}  &\textbf{Loss} &\textbf{Epochs} &\textbf{Batch size} &\textbf{Neurons}\\
\hline
1  & -20.7739 & 75     & 30                             & 80                          \\
2  & -20.7948 & 75     & 60                             & 80                          \\
3  & -20.7988 & 60     & 40                             & 80                          \\
4  & -20.8032 & 60     & 60                             & 110                         \\
5  & -20.8265 & 80     & 50                             & 100                        
\end{tabular}
\end{table}

\begin{table}
\caption{BSA Training Top 5 Performance Table} \label{sample-table}
\centering
\begin{tabular}{lllll}
\textbf{No} & \textbf{MAE} & \textbf{MAPE} & \textbf{Theil's U} \\
\hline
1           & 3.1930       & 337.8235      & 0.6619             \\
2           & 3.0560       & 320.9543      & 0.6485             \\
3           & 3.0869       & 335.1944      & 0.6512             \\
4           & 3.1334       & 352.4468      & 0.6360             \\
5           & 3.1103       & 347.6955      & 0.6490                                   
\end{tabular}
\end{table}

\begin{table}
\caption{BOA Training Top 5 Parameters Table} \label{sample-table1}
\centering
\begin{tabular}{lllll}
\textbf{No} & \textbf{Loss} & \textbf{Epochs} & \multicolumn{1}{l}{\textbf{Batch size}} & \multicolumn{1}{l}{\textbf{Neurons}} \\
\hline
1           & 0.0068        & 70              & 10                                      & 100                                  \\
2           & 0.0432        & 70              & 10                                      & 120                                  \\
3           & 17.9653       & 100             & 10                                      & 70                                   \\
4           & 24.2290       & 70              & 10                                      & 150                                  \\
5           & 51.2767       & 70              & 20                                      & 100                                 
\end{tabular}
\end{table}

\begin{table}
\caption{BOA Training Top 5 Performance Table} \label{sample-table2}
\centering
\begin{tabular}{lllll}
\textbf{No} & \textbf{MAE} & \textbf{MAPE} & \textbf{Theil's U} \\
\hline
1           & 138.3084     & 118.3044      & 0.6734             \\
2           & 152.7400     & 147.7911      & 0.6541             \\
3           & 124.7073     & 107.6635      & 0.6056             \\
4           & 125.4989     & 91.8032       & 0.6172             \\
5           & 123.9566     & 116.2223      & 0.6733                                    
\end{tabular}
\end{table}

\paragraph{SOA} 

\begin{table}
\caption{SOA Training Top 5 Parameters Table} \label{sample-table3}
\centering
\begin{tabular}{lllll}
\textbf{No} & \textbf{Loss} & \textbf{Epochs} & \multicolumn{1}{l}{\textbf{Batch size}} & \multicolumn{1}{l}{\textbf{Neurons}} \\
\hline
1           & 2.9263        & 100             & 30                                      & 175                                  \\
2           & 2.9343        & 100             & 70                                      & 125                                  \\
3           & 2.9738        & 100             & 50                                      & 175                                  \\
4           & 3.0011        & 50              & 70                                      & 125                                  \\
5           & 3.0093        & 75              & 70                                      & 75                                                                 
\end{tabular}
\end{table}

\begin{table}
\caption{SOA Training Top 5 Performance Table} \label{sample-table4}
\centering
\begin{tabular}{lllll}
\textbf{No} & \textbf{MAE} & \textbf{MAPE} & \textbf{Theil's U} \\
\hline
1   & -20.7738  & 75    & 30    & 80    \\ 
2   & -20.7948  & 75    & 60    & 80    \\
3   & -20.7987  & 60    & 40    & 80    \\
4   & -20.8032  & 60    & 60    & 110   \\
5   & -20.8265  & 80    & 50    & 100                                        
\end{tabular}
\end{table}

\begin{table}
\caption{SSA Training Top 5 Parameters Table} \label{sample-table5}
\centering
\begin{tabular}{lllll}
\textbf{No} & \textbf{Loss} & \textbf{Epochs} & \multicolumn{1}{l}{\textbf{Batch size}} & \multicolumn{1}{l}{\textbf{Neurons}} \\
\hline
1           & -20.7738      & 75              & 30                                      & 80                                   \\
2           & -20.7948      & 75              & 60                                      & 80                                   \\
3           & -20.7987      & 60              & 40                                      & 80                                   \\
4           & -20.8032      & 60              & 60                                      & 110                                  \\
5           & -20.8265      & 80              & 50                                      & 100                                                                    
\end{tabular}
\end{table}

\begin{table}
\caption{SSA Training Top 5 Performance Table} \label{sample-table-xxx}
\centering
\begin{tabular}{lllll}
\textbf{No} & \textbf{MAE} & \textbf{MAPE} & \textbf{Theil's U} \\
\hline
1 & 499.8840 & 98.9268       & 0.9870   \\
2 & 500.0705 & 98.9620       & 0.9873   \\
3 & 499.6315 & 98.6651       & 0.9835   \\
4 & 499.5198 & 98.7863       & 0.9852   \\
5 & 500.0753 & 98.9710       & 0.9874              
\end{tabular}
\end{table}

\begin{table}
\caption{Multiagents Train \& Test Results} \label{tab:multiagents1}
\begin{tabular}{lllll}
\centering
\textbf{} & \textbf{Train set} & \textbf{Test set} \\ \hline
Profit per episode(\%) & 0.8070 & 0.3914 \\
Sharpe ratio & 0.4047 & 0.2459 \\
MDD (\%) & -4.2640 & -4.3656 \\
Calmar ratio & 18.9259 & 8.9670        
\end{tabular}
\end{table}

\subsection{Conclusion}
This study focused on short - term forecasts of stock prices. However, Korea's stock market is very disadvantageous to short-term investment because it has a transaction tax of 0.3\% and a transaction fee of 0.03\%. Considering the fact that the volatility of stock prices within a short period of time is not large, it is limited to approach short-term investment in the main Korean stock market by 0.33\% for each transaction. Unlike Korea, the US or Japanese stock market gives transfer tax instead of transaction tax, and transaction tax is 0.1\% in China or Hong Kong, which is much smaller than Korea. Therefore, a rise of 0.33\% is more appropriate for short-term investment because it is profitable even if the first price rises unlike Korea. Therefore, I would like to proceed with the research using market data in the future without transaction tax such as Japan or USA.
In addition, we can consider long-term investment, which is different from short-term investment, which has a strong random-walk attribute, because the stock price follows the value of the company. The value of a company can be assessed quantitatively by annually provided disclosure. In addition, weighing investment through these disclosures over the long term
The fact that it can rise is well proven. Therefore, long-term portfolio composition through multiple disclosures including PER and PBR will be reinforced by learning RL, and further, using Multi-Agent.
First of all, motivation for the start of the study was to validate our assumptions through this experiment and extend it to actual service. Therefore, we plan to develop this model to expand services from short-term recommendation robot advisor service to long-term investment robot adviser service. 

\bibliographystyle{unsrt}
\bibliography{biblio}

\end{document}